\documentclass{article}

\usepackage{arxiv}

\usepackage[utf8]{inputenc} 
\usepackage[T1]{fontenc}    
\usepackage{hyperref}       
\usepackage{url}            
\usepackage{booktabs}       
\usepackage{nicefrac}       
\usepackage{microtype}      
\usepackage{lipsum}		
\usepackage{graphicx}
\usepackage{doi}
\usepackage{algorithm2e}
\RestyleAlgo{ruled}
\usepackage{cite}
\usepackage{amsmath,amssymb,amsfonts}

\title{Multi-view Sparse Laplacian Eigenmaps for nonlinear Spectral Feature Selection}

\author{ \href{https://orcid.org/0000-0003-3139-4439}{\includegraphics[scale=0.1]{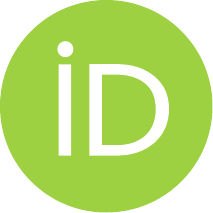}\hspace{1mm}Gaurav Srivastava}\thanks{2023 International Conference on System Science and Engineering (ICSSE 2023). Ho Chi Minh City, Vietnam} \\
	Dept. of Computer Science and Engineering\\
	Manipal University Jaipur\\
	Jaipur, India \\
	\texttt{mailto.gaurav2001@gmail.com} \\
	\And
	\href{https://orcid.org/0000-0001-5697-8786}{\includegraphics[scale=0.1]{orcid.pdf}\hspace{1mm}Mahesh Jangid} \\
	Dept. of Computer Science and Engineering\\
	Manipal University Jaipur\\
	Jaipur, India \\
	\texttt{mahesh\_seelak@yahoo.co.in} \\
}

\date{}

\renewcommand{\shorttitle}{}

\begin{document}
\maketitle

\begin{abstract}
The complexity of high-dimensional datasets presents significant challenges for machine learning models, including overfitting, computational complexity, and difficulties in interpreting results. To address these challenges, it is essential to identify an informative subset of features that captures the essential structure of the data. In this study, the authors propose Multi-view Sparse Laplacian Eigenmaps (MSLE) for feature selection, which effectively combines multiple views of the data, enforces sparsity constraints, and employs a scalable optimization algorithm to identify a subset of features that capture the fundamental data structure. MSLE is a graph-based approach that leverages multiple views of the data to construct a more robust and informative representation of high-dimensional data. The method applies sparse eigendecomposition to reduce the dimensionality of the data, yielding a reduced feature set. The optimization problem is solved using an iterative algorithm alternating between updating the sparse coefficients and the Laplacian graph matrix. The sparse coefficients are updated using a soft-thresholding operator, while the graph Laplacian matrix is updated using the normalized graph Laplacian. To evaluate the performance of the MSLE technique, the authors conducted experiments on the UCI-HAR dataset, which comprises 561 features, and reduced the feature space by 10–90\%. Our results demonstrate that even after reducing the feature space by 90\%, the Support Vector Machine (SVM) maintains an error rate of 2.72\%. Moreover, the authors observe that the SVM exhibits an accuracy of 96.69\% with an 80\% reduction in the overall feature space.
\end{abstract}


\section{Introduction}
In Machine learning (ML), the problem of high dimensionality is widespread, as datasets often contain many variables or features. High dimensionality presents several challenges for ML, including computational complexity, overfitting, and difficulties in interpreting results \cite{b1}. For instance, Activity Recognition (AR), which records accelerometer and gyroscope measurements of human activities, is a prime example of a high-dimensional dataset containing numerous features \cite{b2}. To address this issue, feature selection techniques become crucial in identifying a subset of informative features that capture the essential structure of the data.

One prominent approach to feature selection is spectral Embedding, which maps data into a low-dimensional space by computing the eigenvectors of a Laplacian graph. Laplacian Eigenmaps (LE) is a widely used feature selection method in ML and data analysis. It seeks to preserve the local structure of the data while reducing its dimensionality \cite{b3}. This is achieved by constructing a graph based on pairwise distances between data points and then computing the eigenvectors of the Laplacian matrix associated with this graph \cite{b4}. The Laplacian matrix is a positive semi-definite matrix that encodes the graph's similarity relationships between data points.

However, simple Laplacian eigenmaps are limited by various factors, including scalability issues in high-dimensional data, the inability to handle multiple data views, and a lack of sparsity constraints that could enhance interpretability and computational efficiency. To address these challenges, we present a feature selection method - Multi-view Sparse Laplacian Eigenmaps (MSLE). Our method combines multiple data views, enforces sparsity constraints, and utilizes a scalable optimization algorithm to identify a subset of features that capture the essential data structure. The sparsity constraints encourage the Embedding to use only a small number of features across all views, enhancing interpretability and computational efficiency.

Our approach addresses the issue of high dimensionality by identifying a subset of features common across views and informative for the underlying structure of the data. The sparsity constraints further reduce the dimensionality of the Embedding, resulting in improved interpretability of the results \cite{b5}. The optimization algorithm we employ scales effectively to high-dimensional data and can handle multiple data views, rendering it applicable across a wide range of applications. Our approach's term "Sparse" refers to using sparsity constraints in the optimization objective. Sparsity is a desirable feature in feature selection as it reduces the number of features used in the Embedding, enhancing interpretability and computational efficiency \cite{b6}. In our method, the sparsity constraints are enforced using the $\ell_1$-norm of the embedding matrix, which encourages most entries to be zero.

In this study, we have modified Laplacian Eigenmaps to include sparsity constraints in selecting the eigenvectors. The sparsity constraint promotes selecting a smaller number of eigenvectors that are most relevant to the task at hand. This constraint is enforced through regularization terms in the optimization problem that computes the eigenvectors. Sparse Laplacian Eigenmaps select a smaller set of eigenvectors, resulting in a more interpretable representation of the data. The selected eigenvectors correspond to the most critical features, and their coefficients can be used to interpret the contribution of each feature to the overall representation. Moreover, Sparse Laplacian Eigenmaps select a smaller set of eigenvectors, reducing the computational complexity of the method. This can be especially advantageous for large datasets or when the method needs to be applied in real time. Sparse Laplacian Eigenmaps are less susceptible to data noise and outliers than Laplacian Eigenmaps. The sparsity constraint helps to exclude noisy or irrelevant features, improving the quality of the resulting representation.

The primary contributions of our work are as follows: 

\begin{enumerate}

    \item We present Multi-view Sparse Laplacian Eigenmaps (MSLE) for feature selection, which addresses the challenges posed by high-dimensional datasets. This method utilizes sparse eigendecomposition and an iterative optimization algorithm to construct a robust and informative representation of the data.

    \item We demonstrate the efficacy of our method in enhancing the interpretability and computational efficiency of the Embedding. The results show that MSLE can significantly reduce the feature space while maintaining high classification accuracy, achieving up to a 90\% reduction in the overall feature space.
    
    \item We conduct experiments on the UCI-HAR Activity Recognition benchmark dataset, which consists of 561 attributes.

\end{enumerate}

The remainder of this manuscript is organized as follows. Section II provides background information on Laplacian Eigenmaps and feature selection. Section III details our proposed method, Multi-view Sparse Laplacian Eigenmaps. Section IV reports the results of our experiments, and Section V concludes our work and discusses future directions.

\section{Laplacian Eigenmaps}

Laplacian Eigenmaps is a dimensionality reduction technique that seeks to preserve the local geometric structure of the data \cite{b7}. It does this by constructing a graph from the data points, where each point is a node, and the edges represent the similarity between points.

Let $X = {x_1, x_2, \ldots, x_n}$ be a set of $n$ data points, and let $W$ be an $n \times n$ symmetric weight matrix that represents the pairwise similarity between points. $W(i,j)$ is a measure of the similarity between points $x_i$ and $x_j$ and is typically defined as a function of their Euclidean distance in the feature space.
The Laplacian matrix $L$ is defined as $L = D - W$, where $D$ is a diagonal matrix whose entries are the row sums of $W$. Intuitively, the Laplacian measures the smoothness of the data manifold, and points that are close together in the feature space will have similar values in the Laplacian matrix.

Laplacian Eigenmaps aim to find a low-dimensional embedding $Y = {y_1, y_2, \ldots, y_n}$ of the data points, where each $y_i$ is a $d$-dimensional vector that represents the coordinates of point $x_i$ in the embedded space \cite{b8}. This Embedding is found by solving the following optimization problem:

\begin{equation}
\min _Y \operatorname{Tr}\left(Y^T L Y\right) \quad \text { subject to } Y^T Y=I_d
\end{equation}

where $I_d$ is the $d \times d$ identity matrix, and $\mathrm{Tr}$ denotes the trace of a matrix. The solution to this problem is given by the eigenvectors corresponding to the $d$ smallest eigenvalues of the generalized eigenvalue problem:

\begin{equation}
L v=\lambda D v
\end{equation}

where $\lambda$ is the eigenvalue, $v$ is the eigenvector, and $D$ is the diagonal matrix defined above.
The Laplacian Eigenmaps method can be extended to handle multiple views of the data by constructing a separate Laplacian matrix for each view and then integrating them using a weighted combination. This is known as Multi-view Laplacian Eigenmaps.

\subsection{Spectral Embedding for Non-Linear Dimensionality Reduction}
Spectral Embedding is a family of algorithms for non-linear dimensionality reduction based on the spectral decomposition of a graph Laplacian \cite{b9}. In this approach, the data is first transformed into a graph by computing pairwise similarities between data points. This graph is then represented by its Laplacian matrix, which captures the graph's topology.

Let $X = {x_1, x_2, \ldots, x_n}$ be a set of $n$ data points, and let $W$ be an $n \times n$ symmetric affinity matrix that represents the pairwise similarities between points. The affinity matrix can be defined in various ways depending on the application. For example, it could be a Gaussian kernel function of the Euclidean distances between points:

\begin{equation}
W(i, j)=e^{-\frac{\left\|x_i-x_j\right\|^2}{2 \sigma^2}}
\end{equation}

where $\sigma$ is a parameter that controls the width of the kernel.

The Laplacian matrix $L$ is then defined as $L = D - W$, where $D$ is a diagonal matrix whose entries are the row sums of $W$. This matrix captures the geometry of the graph and is used to perform spectral decomposition.

The goal of spectral Embedding is to find a low-dimensional representation of the data points that preserves the geometric structure of the graph. This is achieved by computing the eigenvectors of the Laplacian matrix corresponding to the $d$ smallest eigenvalues and using them to define the coordinates of the embedded points. Specifically, let $v_i$ be the $i$-th eigenvector of $L$ corresponding to the $i$-th smallest eigenvalue $\lambda_i$. The coordinates of the $j$-th data point in the embedded space are given by:

\begin{equation}
y_j=\left(v_1(j), v_2(j), \ldots, v_d(j)\right)
\end{equation}

where $v_i(j)$ denotes the $j$-th element of the $i$-th eigenvector.
The Laplacian matrix is normalized to ensure that the Embedding is well-scaled. Two standard normalization methods are symmetric normalization and random walk normalization, which are defined as:

\begin{equation}
\begin{gathered}
\operatorname{Sym}(L)=D^{-1 / 2} L D^{-1 / 2} \\
\mathrm{RW}(L)=D^{-1} L
\end{gathered}
\end{equation}

respectively, where $D^{-1/2}$ and $D^{-1}$ are the diagonal matrices with the reciprocal square roots and the reciprocals of the diagonal entries of $D$, respectively.

\section{Proposed Method - Multi-view Sparse Laplacian Eigenmaps}

Given a dataset $X = [x_1, x_2, ..., x_n] \in \mathbb{R}^{d\times n}$ with $n$ samples and $d$ features, we seek to identify a subset of $k$ features that capture the essential structure of the data. Our approach, Multi-view Sparse Laplacian Eigenmaps (MSLE), combines multiple data views and enforces sparsity constraints to identify the most informative features. We construct a graph $G=(V, E)$, where $V$ is the set of vertices corresponding to the $n$ samples, and $E$ is the edges representing pairwise relationships between samples.

We begin by computing the graph Laplacian $L$ associated with the graph $G$, defined as $L=D-W$, where $D$ is the degree matrix and $W$ is the weighted adjacency matrix. The $i$th diagonal element of $D$ is defined as $D_{ii} = \sum_{j=1}^{n} W_{ij}$, and the $(i,j)$th element of $W$ is the similarity between samples $i$ and $j$, e.g., the Gaussian kernel $W_{ij} = \exp(-\frac{1}{2\sigma^2} ||x_i-x_j||^2)$. We then compute the eigenvectors $U_k = [u_1,u_2,...,u_k]$ and eigenvalues $\lambda_k = [\lambda_1,\lambda_2,...,\lambda_k]$ of $L$.

To enforce sparsity constraints, we add an $\ell_1$-norm penalty term to the objective function, resulting in the following optimization problem:

\begin{equation}
\min_{Z}\left(\sum_{i=1}^k \lambda_i \left|x-U_kz_i\right|^2 + \alpha\left|z_i\right|_1\right)
\end{equation}

where $z_i$ is the $i$th column of $Z\in\mathbb{R}^{k\times n}$, and $\alpha$ is a hyperparameter that controls the degree of sparsity. The first term minimizes the reconstruction error between the original data and the low-dimensional Embedding, while the second term encourages sparsity in the Embedding by penalizing nonzero entries.

We use the accelerated proximal gradient method with a backtracking line search to solve the optimization problem. This method iteratively updates the variable $Z$ as follows:

\begin{equation}
\begin{split}
Z^{(t)} = \text{argmin}{Z} &\left(\sum{i=1}^{k} \lambda_i \left|x-U_kz_i\right|^2 \right.
\\
&\left.+\alpha\left|z_i^{(t-1)} + \frac{1}{\beta}(U_k^T(x-U_kz_i^{(t-1)}))\right|_1\right)
\end{split}
\end{equation}

where $t$ is the iteration number and $\beta$ is the Lipschitz constant of the gradient of the $\ell_1$-norm penalty term. This method has a convergence rate of $\mathcal{O}(1/t^2)$.

Once we obtain the embedding matrix $Z$, we select the $k$ most informative features by thresholding the absolute values of the entries in $z_i$. The selected features correspond to the columns of $U_k$ with the largest absolute values in the corresponding rows of $Z$.

The goal of Multi-view Sparse Laplacian Eigenmaps is to find a low-dimensional representation of the data that preserves the structure of the combined Laplacian matrix while selecting a subset of features that are common across views and informative for the underlying structure \cite{b10}. This is achieved by adding sparsity constraints to the Laplacian Eigenmaps objective function, encouraging the Embedding to use only a few features \cite{b11}. Specifically, the objective function is given by:
\begin{equation}
\operatorname{minimize} \quad \operatorname{trace}\left(Y^T L Y\right)+\lambda\|Y\|_1^1
\end{equation}
where $Y$ is the $n \times d$ embedding matrix, $d$ is the dimensionality of the embedded space, $\lambda$ is a sparsity parameter, and $||\cdot||_1^1$ denotes the $\ell_1$-norm of the matrix. The sparsity constraint encourages Embedding to use only a small number of features across all views.

\subsection{Algorithm}
This algorithm \ref{alg:multi_view_sparse_laplacian_eigenmaps} computes the multi-view Laplacian matrix by summing the Laplacian matrices of each view, and computes a sparse weight matrix that maximizes the agreement between the views. It then computes the eigenvectors and eigenvalues of the multi-view Laplacian matrix and selects the top $k$ eigenvectors as the selected features. The algorithm returns both the weight matrix and the selected features. The algorithm includes regularization parameters $\alpha_i$ for each view, which can be chosen using cross-validation.

\begin{algorithm}[htbp]
\caption{Multi-view Sparse Laplacian Eigenmaps}
\label{alg:multi_view_sparse_laplacian_eigenmaps}
\KwIn{
${X_1, X_2, \ldots, X_m}$: $m$ views of the dataset,
$k$: number of selected features,

$\{\alpha_1, \alpha_2, \ldots, \alpha_m\}$: regularization parameters.
}
\vspace{3mm}

\KwOut{
$W$: weight matrix,
$F$: selected features.
}
\vspace{3mm}

\textbf{1:} Compute the pairwise similarity matrices $S_i$ for each view $i$.
\vspace{1mm}

\textbf{2:} Compute the Laplacian matrices $L_i$ for each view $i$.
\vspace{1mm}

\textbf{3:} Compute the diagonal matrices $D_i$ for each view $i$.
\vspace{1mm}

\textbf{4:} Compute the multi-view Laplacian matrix $L$ as $L = \sum_{i=1}^m L_i$.
\vspace{1mm}

\textbf{5:} Compute the multi-view diagonal matrix $D$ as $D = \sum_{i=1}^m D_i$.
\vspace{1mm}

\textbf{6:} Compute the sparse weight matrix $W$ by solving the following optimization problem:
\begin{equation*}
\min_W \sum_{i=1}^m |X_i - X_i W|F^2 + \sum_{i=1}^m \alpha_i \mathrm{Tr}(W^T L_i W)
\end{equation*}

\textbf{7:} Compute the eigenvectors $U$ and eigenvalues $\Lambda$ of the generalized eigenvalue problem $L U = D U \Lambda$.
\vspace{1mm}

\textbf{8:} Select the top $k$ eigenvectors corresponding to the $k$ smallest eigenvalues and stack them into the matrix $F$.

\vspace{3mm}
\textbf{9:} \textbf{Return} $W$ and $F$.
\end{algorithm}

\section{Experimental Results and Discussion}

\subsection{Dataset Used: UCI-HAR}

The UCI-HAR dataset comprises recordings of 30 individuals carrying a smartphone fitted with inertial sensors on their waist, as they engaged in their daily activities \cite{b12}. The participants, aged between 19 and 48 years, were instructed to perform six activities, including walking, sitting, and lying down, while holding a Samsung Galaxy S II smartphone. The smartphone's accelerometer and gyroscope recorded 3-axial linear acceleration and 3-axial angular velocity at a fixed sampling rate of 50Hz. The activities were manually annotated using video recordings \cite{b13}, and the resulting dataset was randomly partitioned into training and test sets, with 70\% of the participants assigned to the training set as shown in table \ref{table: data1}.


\begin{table}[htbp]
\centering
\caption{Description of the UCI-HAR dataset, including the number of training and testing samples available for each activity.}

\label{table: data1}
\begin{tabular}{|c|c|c|c|}
\hline
\textbf{Activity} & \textbf{\begin{tabular}[c]{@{}l@{}}Total no. of\\ Samples\end{tabular}} & \textbf{\begin{tabular}[c]{@{}l@{}}Training\\ Samples\end{tabular}} & \textbf{\begin{tabular}[c]{@{}l@{}}Testing\\ Samples\end{tabular}} \\
\hline
LAYING & 1944 & 1407 & 537 \\
STANDING & 1906 & 1374 & 532 \\
SITTING & 1777 & 1286 & 496 \\
WALKING & 1722 & 1226 & 491 \\
WALKING\_UPSTAIRS & 1544 & 1073 & 471 \\
WALKING\_DOWNSTAIRS & 1406 & 986 & 420 \\
\hline
 & \textbf{10299} & \textbf{7352} & \textbf{2947} \\
\hline
\end{tabular}
\end{table}

\subsection{Experimental Settings}
The experiments in this study were carried out using a 12th Gen Intel(R) Core(TM) i7-1265U processor with 32 GB of memory.

\subsection{t-Distributed Stochastic Neighbor Embedding (t-SNE)}
\begin{figure}[htbp]
\centering
\includegraphics[width=0.6\textwidth]{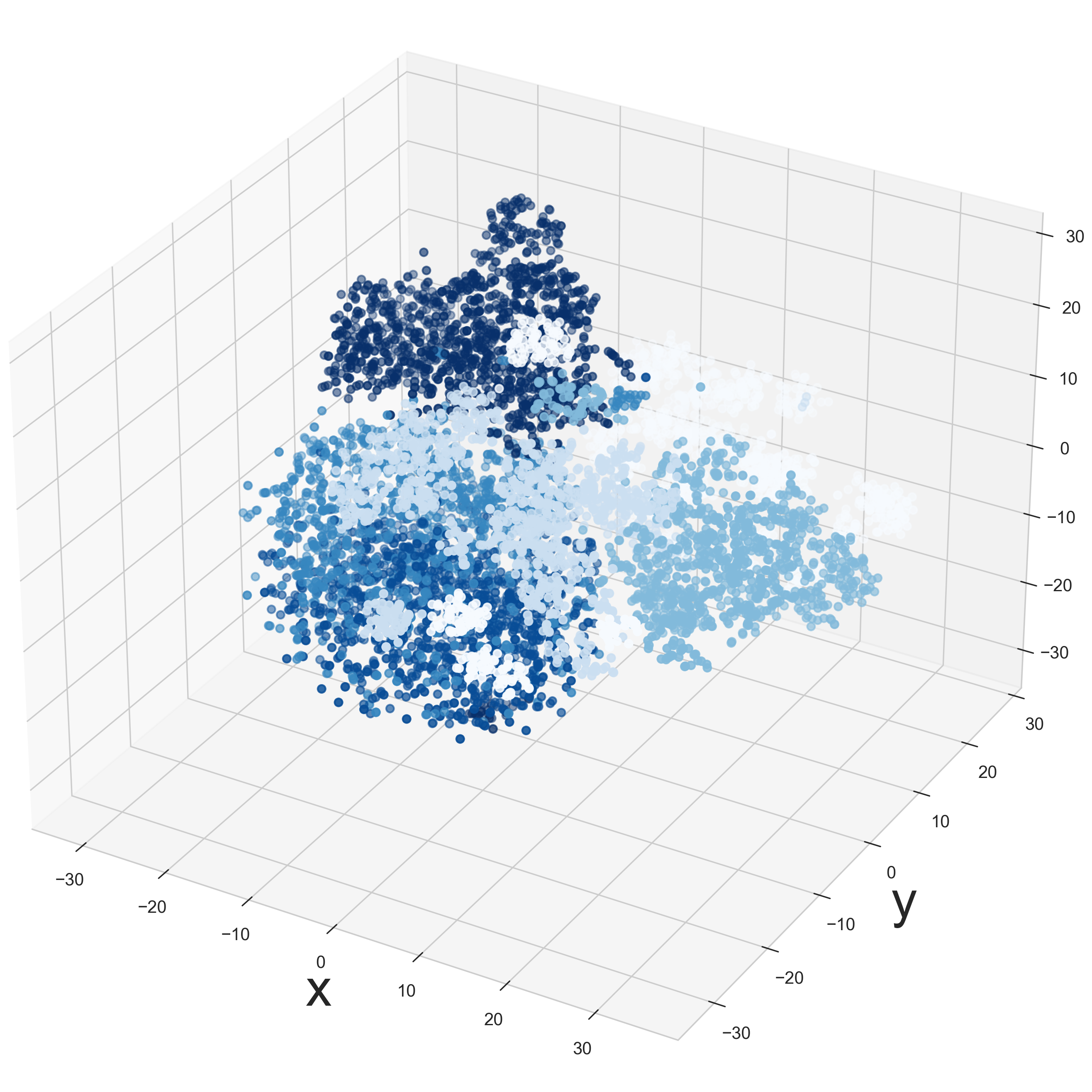}
\caption{$3$ dimensional view of the UCI-HAR dataset.}
\label{fig: 3d}
\end{figure}

To obtain a more comprehensive understanding of the high-dimensional time series data used in our study, we applied the t-SNE method \cite{b14}. As depicted in figure \ref{fig: 3d}, the complexity of the data in its original form posed a challenge to comprehend its underlying structure.

\begin{figure}[htbp]
\centering
\includegraphics[width=0.6\textwidth]{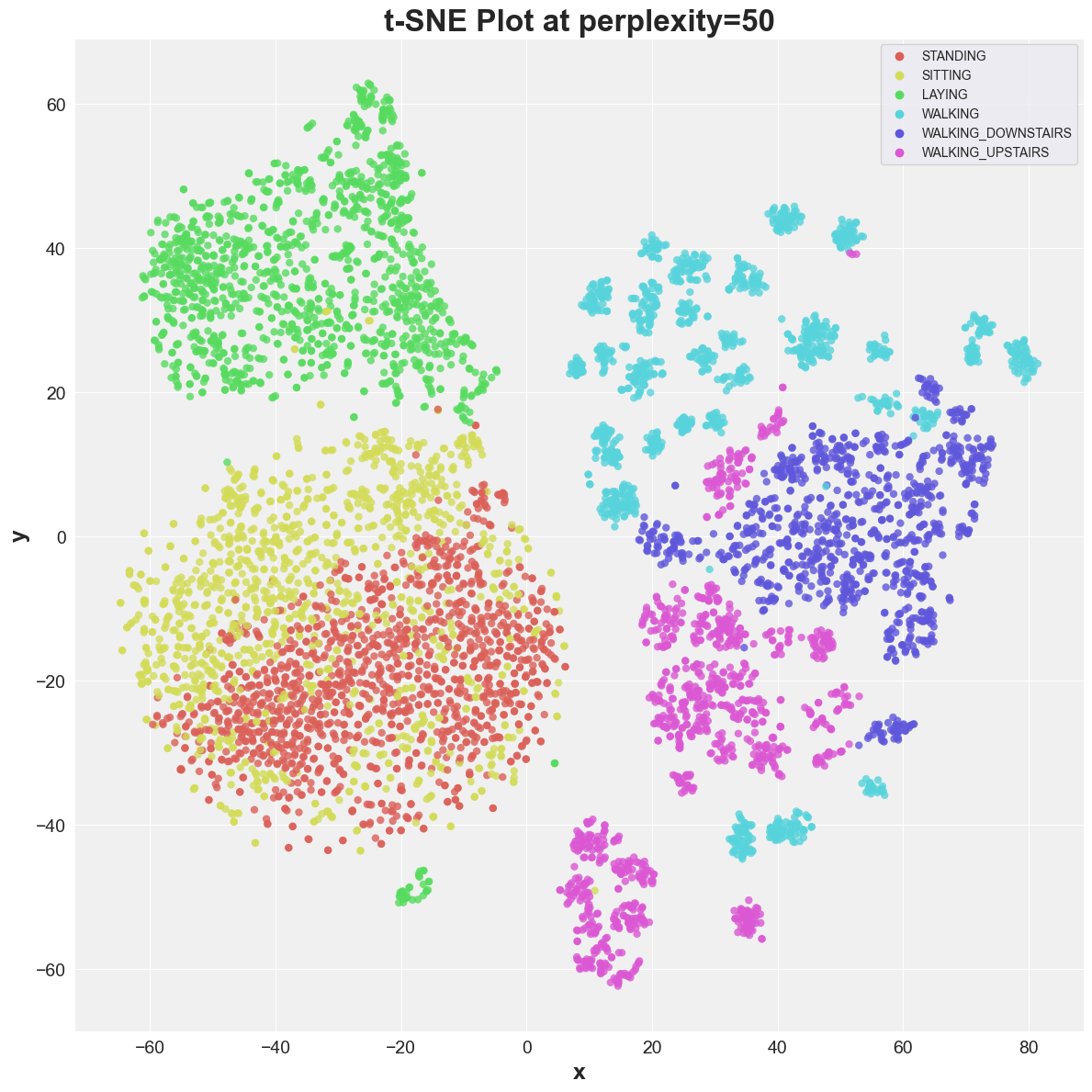}
\caption{Visualization of high-dimensional data in low dimension using the t-SNE algorithm on the UCI-HAR dataset.}
\label{fig: tsne}
\end{figure}

\begin{figure}[htbp]
\centering
\includegraphics[width=0.5\textwidth]{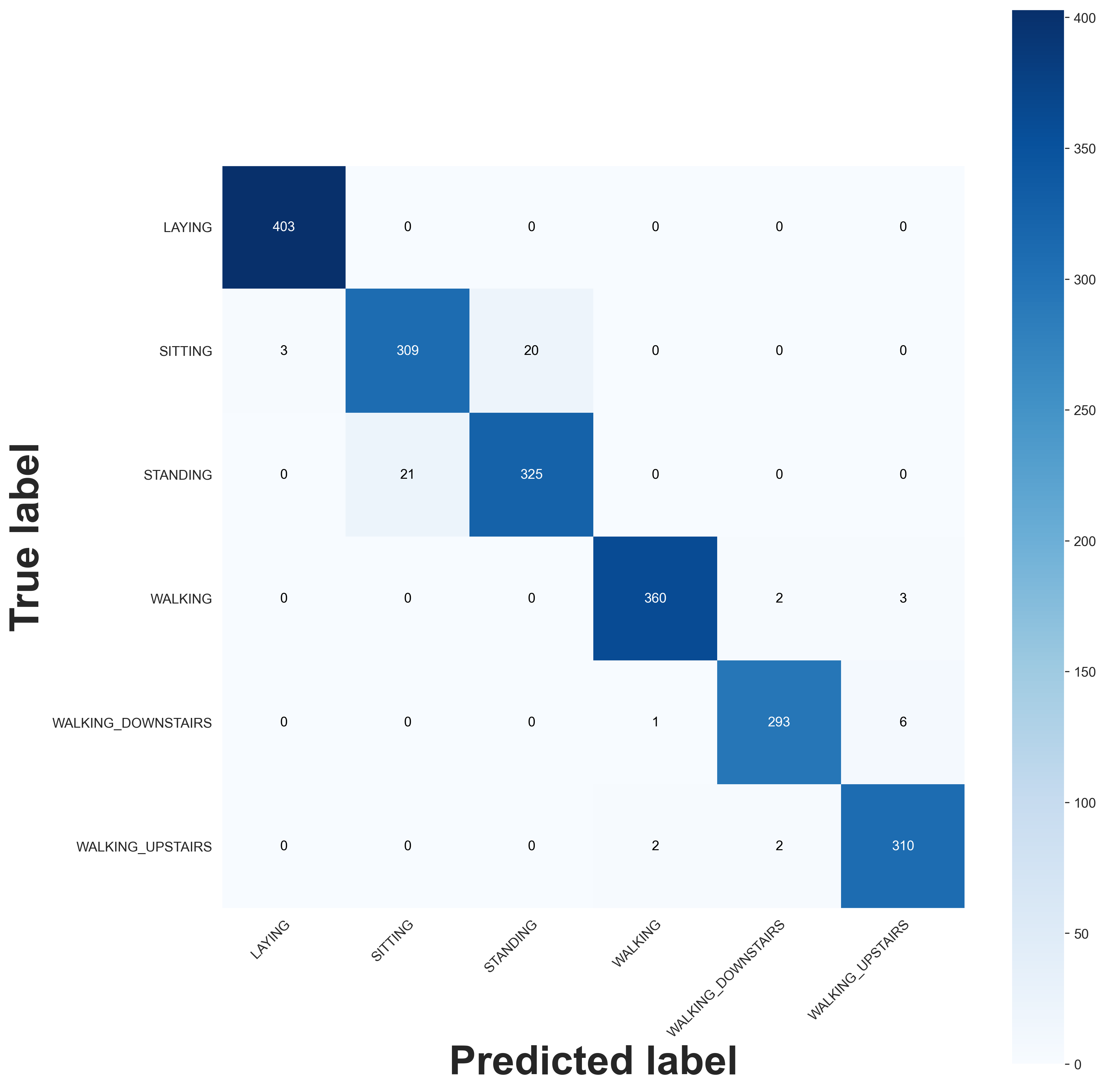}
\caption{Confusion Matrix for the trained SVM on 80\% features.}
\label{fig: CF}
\end{figure}

The t-SNE algorithm was employed on the UCI-HAR data, with perplexity values ranging from 5 to 50, for 1000 iterations, as shown in figure \ref{fig: tsne}. The visualization revealed that at a perplexity of 50, all features except STANDING and SITTING were readily separable. Thus, the model may be confused between the standing and sitting classes, leading to prediction errors. Additionally, some WALKING features overlapped with WALKING\_UPSTAIRS features, indicating the potential for prediction inaccuracies in this area. Subsequently, figure \ref{fig: CF} will provide further visual evidence to support these observations. By utilizing t-SNE and considering the perplexity value, valuable insights into the data's characteristics can be obtained, which can inform the selection of an appropriate machine learning algorithm and enhance the model's performance \cite{b15}.

\begin{table}[htbp]
\caption{Reported Metrics of 12 distinct machine learning classifiers when trained on UCI-HAR dataset.}
	\label{table: classifiers}
\centering
\begin{tabular}{|c|c|c|c|c|}
\hline
\textbf{Classifier} & \textbf{Accuracy} & \textbf{Precision} & \textbf{Recall} & \textbf{F1-Score} \\
\hline
KNN & 80.9 & 81.8 & 81.0 & 81.0 \\
Gaussian Naive Bayes & 77.0 & 79.3 & 77.0 & 76.8 \\
Decision Trees & 85.2 & 85.3 & 85.2 & 85.2 \\
Random Forest & 92.0 & 92.2 & 92.1 & 92.1 \\
Extra Trees & 94.2 & 94.4 & 94.3 & 94.2 \\
SVM & 93.0 & 93.3 & 93.1 & 93.0 \\
Multi-layer Perceptron & 93.8 & 94.4 & 93.9 & 93.9 \\
XGBoost & 93.4 & 93.5 & 93.4 & 93.4 \\
LightGBM & 93.1 & 93.3 & 93.1 & 93.1 \\
CatBoost & 92.7 & 92.9 & 92.8 & 92.8 \\
LDA  & 96.43 & 96.5 & 96.4 & 96.4 \\
QDA & 70.4 & 80.0 & 70.4 & 69.8\\
\hline
\end{tabular}
\end{table}

\begin{table*}[htbp]
\caption{Test accuracies of 12 distinct machine learning classifiers when trained on reduced feature subsets of range 10\% -- 90\%.}
\label{table: 2}
\centering
\begin{tabular}{|c|c|c|c|c|c|c|c|c|c|}
\hline
\textbf{Classifier} & \textbf{10\%} & \textbf{20\%} & \textbf{30\%} & \textbf{40\%} & \textbf{50\%} & \textbf{60\%} & \textbf{70\%} & \textbf{80\%} & \textbf{90\%} \\
\hline
KNN & 68.44 & 74.90 & 79.32 & 83.93 & 86.84 & 90.19 & 90.43 & 92.81 & 91.94 \\

Gaussian Naive Bayes & 56.06 & 62.76 & 63.05 & 64.02 & 66.69 & 71.79 & 76.45 & 81.35 & 80.87 \\

Decision Trees & 83.54 & 82.76 & 84.75 & 83.73 & 84.07 & 84.22 & 84.27 & 85.09 & 85.38 \\
Random Forest & 91.89 & 92.08 & 91.50 & 91.89 & 92.57 & 92.37 & 92.37 & 93.59 & 91.31 \\
Extra Trees & 91.60 & 91.74 & 91.65 & 92.86 & 93.30 & 92.86 & 92.66 & 94.17 & 91.65 \\
SVM & 96.74 & 97.08 & 96.60 & 97.13 & 97.23 & 97.18 & 96.50 & 96.69 & 94.02 \\
Multi-layer Perceptron & 91.16 & 91.50 & 92.28 & 92.18 & 92.42 & 92.52 & 91.94 & 89.85 & 87.96 \\
XGBoost & 95 & 95.24 & 95.19 & 94.56 & 95.04 & 95.97 & 95.19 & 95.38 & 93.68 \\
LightGBM & 94.32 & 95.04 & 94.95 & 94.70 & 95.48 & 95.38 & 94.80 & 95.33 & 93.54 \\
CatBoost & 95.29 & 95.63 & 95.14 & 95.48 & 95.38 & 95.77 & 95.19 & 95.24 & 93.39 \\
LDA & 94.70 & 95.43 & 94.56 & 95.29 & 94.95 & 94.90 & 93.49 & 93.30 & 89.07 \\
QDA & 83.83 & 82.96 & 83.20 & 83.93 & 84.95 & 84.17 & 83.83 & 85.67 & 83.44\\
\hline
\end{tabular}
\end{table*}

\subsection{Classification Performance and MSLE Analysis}
To assess the efficacy of the suggested approach, we employed MSLE based feature selection method on the UCI-HAR dataset. We conducted experiments to assess the performance of various classifiers with reduced feature dimensions. Initially, we applied 13 machine learning classifiers to the complete feature set to obtain a general idea of their performance on time-series data. Linear Discriminant Analysis (LDA) emerged as the top performer, achieving an accuracy of 96.43\% on the UCI-HAR dataset. Support Vector Machines (SVM) and Boosting models also demonstrated promising results. The accuracy, precision, recall, and f1-score of all the evaluated classifiers are presented in table \ref{table: classifiers}.

Subsequently, we progressively reduced the feature sets using MSLE and assessed the robustness of the method. The results in table \ref{table: 2} indicate that the accuracy of the classifiers remained relatively stable as the feature dimensions were reduced from 10\% to 90\%, with some classifiers even demonstrating increased accuracy. This increase can be attributed to the fact that not all features contribute equally to the classification task, and some features may even be irrelevant or redundant. By reducing the number of features, the model can focus on the most informative and relevant ones, improving classification performance.

However, it is important to note that reducing the feature dimensions excessively can lead to underfitting, where the model is too simplistic and fails to capture the complexity of the data. Therefore, it is crucial to strike a balance between feature reduction and model complexity and to carefully evaluate the model's performance at each stage of the feature selection process. Reducing feature dimensions by 40-50\% can be a suitable approach to mitigate the high dimensionality problem while ensuring computational efficiency.

\begin{figure}[htbp]
\centering
\includegraphics[width=0.5\textwidth]{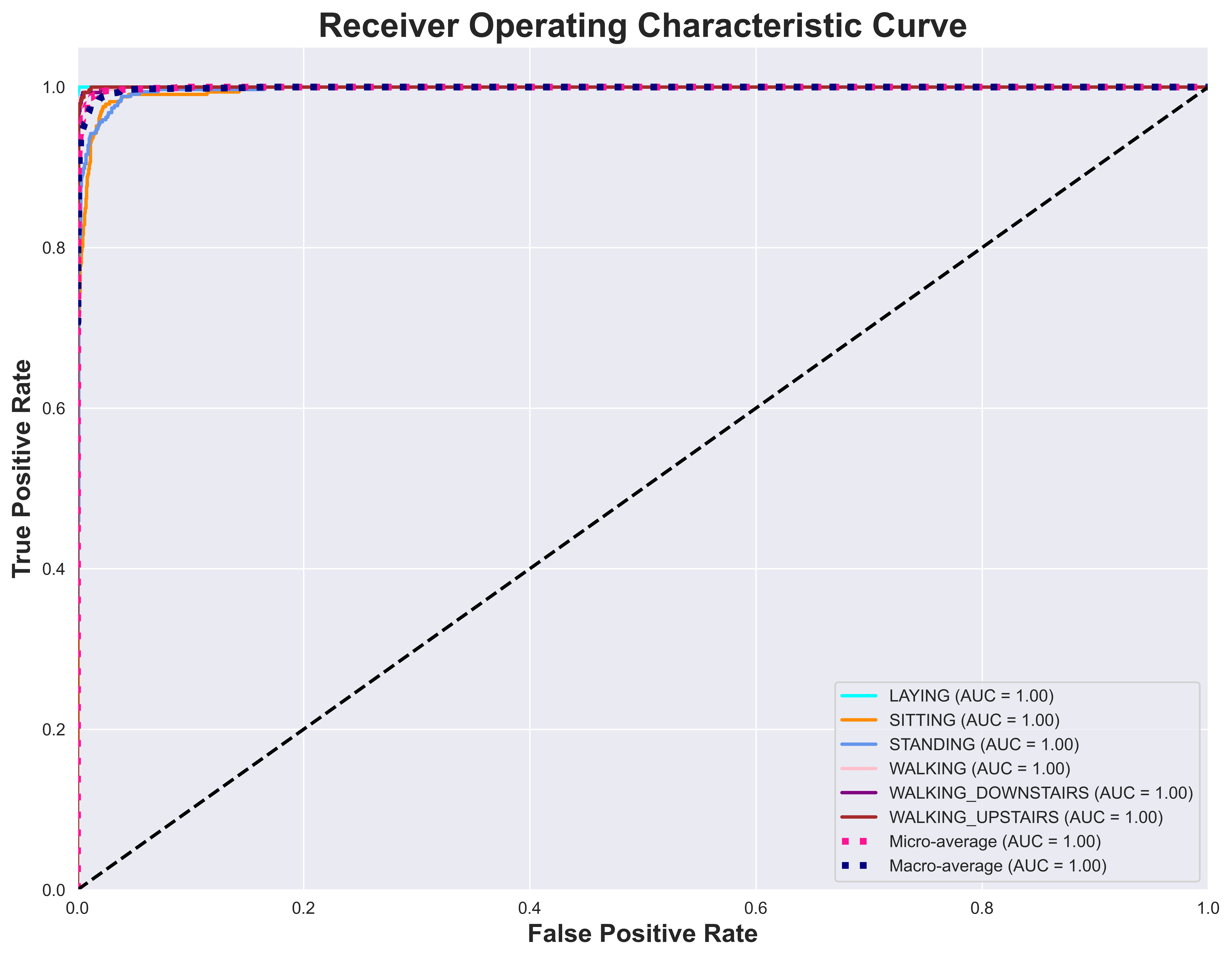}
\caption{Receiver Operating Characteristic curve for the trained SVM on 80\% features.}
\label{fig: ROC}
\end{figure}

For instance, the K-Nearest Neighbors (KNN) classifier achieved an accuracy of 68.44\% with 10\% of the features reduced, which increased to 91.94\% with 90\% of the features reduced. However, it is possible that the model may be underfitting with such a high feature reduction. Conversely, some classifiers, such as the SVM and XGBoost, demonstrated consistently high accuracy across all feature dimensions. The SVM classifier achieved an accuracy of 96.74\% with 10\% of the features, which peaked at 97.23\% with 50\% of the features before declining to 94.02\% with 90\% of the features as shown in table \ref{table: 2}. Similarly, the XGBoost classifier achieved an accuracy of 95.00\% with 10\% of the features, which peaked at 95.97\% with 50\% of the features before declining to 93.68\% with 90\% of the features. The confusion matrix and ROC plots of the trained SVM with an 80\% feature reduction are depicted in figures \ref{fig: CF} and \ref{fig: ROC}, respectively.

The results demonstrate the effectiveness of the MSLE technique in reducing the dimensionality of the UCI-HAR dataset while maintaining or improving the classification accuracy of various classifiers. 

\subsection{Analysis on Computational Efforts}

To reduce the dimensionality of the data, MSLE applies sparse eigendecomposition to the Laplacian matrix, which yields a small set of eigenvectors that capture the most important and informative features of the data. These eigenvectors serve as the reduced feature set, which can be used for classification tasks. In our experiments, MSLE has demonstrated efficiency and scalability, making it a promising technique for real-world scenarios where processing large amounts of high-dimensional data. The runtime of the MSLE technique is dependent on the size of the dataset and the number of views utilized to construct the graph. In the experimental analysis conducted on the UCI-HAR dataset, MSLE took an average of 6 minutes and 13 seconds to reduce the feature dimensions using all six data views. However, the runtime can be further optimized by decreasing the number of views or employing parallel computing techniques.

\section{Conclusion}
Dealing with high-dimensional data presents challenges in modeling and may lead to sparsity issues. In this study, we present Multi-view Sparse Laplacian Eigenmaps (MSLE) for feature selection, which addresses the challenges presented by high-dimensional datasets. We demonstrate the effectiveness of the MSLE technique through experiments on the UCI-HAR dataset, where it significantly reduces the feature space while maintaining high classification accuracy, achieving up to an 80\% reduction in the overall feature space.

MSLE utilizes multiple views of the data, sparse eigendecomposition, and an iterative optimization algorithm to construct a robust and informative representation of high-dimensional data. These features make MSLE a promising approach for practical applications where high-dimensional datasets are prevalent. The results of this study have significant implications for the field of machine learning, as they demonstrate the potential of MSLE to overcome challenges posed by high-dimensional datasets, such as overfitting and computational complexity. Additionally, MSLE has the potential to contribute to the development of more efficient and interpretable machine learning models.

Future research could investigate the effectiveness of MSLE on other high-dimensional datasets and the potential of integrating it with other feature selection techniques to further improve classification performance. Furthermore, the applicability of MSLE to other domains, such as social network analysis and natural language processing, could be explored. Evaluating the performance of MSLE on datasets with different data types and structures could also provide insights into its versatility and effectiveness. Finally, the development of new optimization algorithms and techniques could enhance the scalability of MSLE for processing large datasets.

\section*{Acknowledgment}
The authors would like to express our gratitude to the Department of Computer Science and Engineering at Manipal University Jaipur for their valuable support and encouragement, as well as for providing us with all the necessary resources to carry out our experiment successfully.







\end{document}